\documentclass[11pt]{article}
\usepackage{coling2018}
\usepackage{times}
\usepackage{url}
\usepackage{latexsym}
\usepackage{pifont}
\usepackage{array}
\usepackage{graphicx}
\usepackage{floatrow}
\newfloatcommand{capbtabbox}{table}[][\FBwidth]

\newcommand{\cmark}{\ding{51}}

\newcolumntype{L}{>{\centering\arraybackslash}p{3.5cm}}
\newcolumntype{K}{>{\centering\arraybackslash}p{1.5cm}}

\newcolumntype{G}{>{\centering\arraybackslash}p{0.9cm}}
\newcolumntype{H}{>{\centering\arraybackslash}p{1.8cm}}
\newcolumntype{M}{>{\centering\arraybackslash}p{1.1cm}}
\newcolumntype{D}{>{\centering\arraybackslash}p{2.2cm}}
\newcolumntype{C}{>{\centering\arraybackslash}p{0.9cm}}
\newcolumntype{B}{>{\centering\arraybackslash}p{2cm}}
\newcolumntype{A}{>{\centering\arraybackslash}p{0.8cm}}
\newcolumntype{Z}{>{\centering\arraybackslash}p{1.1cm}}
\newcolumntype{T}{>{\centering\arraybackslash}p{1.2cm}}
\newcolumntype{V}{>{\centering\arraybackslash}p{1.4cm}}

\newcolumntype{U}{>{\centering\arraybackslash}p{0.8cm}}

\title{Bringing replication and reproduction together with generalisability in NLP: Three reproduction studies for Target Dependent Sentiment Analysis}
\author{Andrew Moore and Paul Rayson\\
   School of Computing and Communications, Lancaster University, Lancaster, UK \\
  {\tt initial.surname@lancaster.ac.uk}}

\date{}
\begin{document}
\maketitle
\begin{abstract}
Lack of repeatability and generalisability are two significant threats to continuing scientific development in Natural Language Processing. Language models and learning methods are so complex that scientific conference papers no longer contain enough space for the technical depth required for replication or reproduction. Taking Target Dependent Sentiment Analysis as a case study, we show how recent work in the field has not consistently released code, or described settings for learning methods in enough detail, and lacks comparability and generalisability in train, test or validation data. To investigate generalisability and to enable state of the art comparative evaluations, we carry out the first reproduction studies of three groups of complementary methods and perform the first large-scale mass evaluation on six different English datasets. Reflecting on our experiences, we recommend that future replication or reproduction experiments should always consider a variety of datasets alongside documenting and releasing their methods and published code in order to minimise the barriers to both repeatability and generalisability. We have released our code with a model zoo on GitHub with Jupyter Notebooks to aid understanding and full documentation, and we recommend that others do the same with their papers at submission time through an anonymised GitHub account. 

\end{abstract}

\section{Introduction}
%
%
\blfootnote{
    %
    %
    %
    %
     \hspace{-0.65cm}  
     This work is licenced under a Creative Commons 
     Attribution 4.0 International Licence.
     Licence details:
     \url{http://creativecommons.org/licenses/by/4.0/}
    %
    %
}

Repeatable (replicable and/or reproducible\footnote{We follow the definitions in Antske Fokkens' guest blog post ``replication (obtaining the same results using the same experiment) as well as reproduction (reach the same conclusion through different means)'' from \url{http://coling2018.org/slowly-growing-offspring-zigglebottom-anno-2017-guest-post/}}) experimentation is a core tenet of the scientific endeavour. In Natural Language Processing (NLP) research as in other areas, this requires three crucial components: (a) published methods described in sufficient detail (b) a working code base and (c) open dataset(s) to permit training, testing and validation to be reproduced and generalised. In the cognate sub-discipline of corpus linguistics, releasing textual datasets has been a defining feature of the community for many years, enabling multiple comparative experiments to be conducted on a stable basis since the core underlying corpora are community resources. In NLP, with methods becoming increasingly complex with the use of machine learning and deep learning approaches, it is often difficult to describe all settings and configurations in enough detail without releasing code. The work described in this paper emerged from recent efforts  
at our research centre to reimplement other's work across a number of topics (e.g. text reuse, identity resolution and sentiment analysis) where previously published methods were not easily repeatable because of missing or broken code or dependencies, and/or where methods were not sufficiently well described to enable reproduction. We focus on one sub-area of sentiment analysis to illustrate the extent of these problems, along with our initial recommendations and contributions to address the issues.

The area of Target Dependent Sentiment Analysis (TDSA) and NLP in general has been growing rapidly in the last few years due to new neural network methods that require no feature engineering. However it is difficult to keep track of the state of the art as new models are tested on different datasets, thus preventing true comparative evaluations. This is best shown by table \ref{table:methods_datasets} where many approaches are evaluated on the SemEval dataset \cite{pontiki_2014} but not all. Datasets can vary by domain (e.g. product), type (social media, review), or medium (written or spoken), and to date there has been no comparative evaluation of methods from these multiple classes. Our primary and secondary contributions therefore, are to carry out the first study that reports results across all three different dataset classes, and to release a open source code framework implementing three complementary groups of TDSA methods.

In terms of reproducibility via code release, recent TDSA papers have generally been very good with regards to publishing code alongside their papers \cite{mitchell_2013,zhang_2015,zhang_2016,liu_2017,Marrese_Taylor_2017,wang_2017} but other papers have not released code \cite{wang_2016,tay_2017}. In some cases, the code was initially made available, then removed, and is now back online \cite{Tang_2016_tdlstm}. Unfortunately, in some cases even when code has been published, different results have been obtained relative to the original paper. This can be seen when \newcite{chen_2017} used the code and embeddings in \newcite{Tang_2016_mem} they observe different results. Similarly, when others \cite{tay_2017,chen_2017} attempt to replicate the experiments of \newcite{Tang_2016_tdlstm} they also produce different results to the original authors. Our observations within this one sub-field motivates the need to investigate further and understand how such problems can be avoided in the future. In some cases, when code has been released, it is difficult to use which could explain why the results were not reproduced. Of course, we would not expect researchers to produce industrial strength code, or provide continuing free ongoing support for multiple years after publication, but the situation is clearly problematic for the development of the new field in general.

In this paper, we therefore reproduce three papers 
chosen as they employ widely differing methods: Neural Pooling (NP) \cite{vo_2015}, NP with dependency parsing \cite{wang_2017}, and RNN \cite{Tang_2016_tdlstm}, as well as having been applied largely to different datasets. At the end of the paper, we reflect on bringing together elements of repeatability and generalisability which we find are crucial to NLP and data science based disciplines more widely to enable others to make use of the science created.
\begin{table}[!h]
\centering
\begin{tabular}{|l|M|M|M|M|M|M|}
\hline
 &  \multicolumn{6}{c|}{Datasets}  \\
\hline
Methods & 1 & 2 & 3 & 4 & 5 & 6\\
 \hline
 \hline
 \newcite{mitchell_2013} &  &  & \cmark   & & & \\
 \hline
 \newcite{kiritchenko_2014} &  &  &  & \cmark &  &  \\
 \hline
 \newcite{dong_2014} & \cmark &  & & & & \\
 \hline
 \newcite{vo_2015} & \cmark & \cmark & \cmark  & & &  \\
 \hline
 \newcite{zhang_2015} & &  & \cmark  & & & \\
 \hline
 \newcite{zhang_2016} & \cmark & \cmark & \cmark  & & &  \\
 \hline
 \newcite{Tang_2016_tdlstm} & \cmark &   &  & \cmark & & \\
 \hline
 \newcite{Tang_2016_mem} & & &   & \cmark & & \\
 \hline
 \newcite{wang_2016} & &  & & \cmark & & \\
 \hline
 \newcite{chen_2017} & \cmark &  &   & \cmark & & \\
 \hline
 \newcite{liu_2017} & \cmark & \cmark & \cmark  &  & &  \\
 \hline
 \newcite{wang_2017} & \cmark & &   &  & \cmark &  \\
 \hline
 \newcite{Marrese_Taylor_2017} & &  & &   \cmark & &  \cmark\\
 \hline
 \hline
\multicolumn{7}{|p{13.5cm}|}{1=\newcite{dong_2014}, 2=\newcite{wilson_2008}, 3=\newcite{mitchell_2013}, 4=\newcite{pontiki_2014}, 5=\newcite{wang_2017}, 6=\newcite{Marrese_Taylor_2017}}\\
\hline
\end{tabular}
\caption{Methods and Datasets}
\label{table:methods_datasets}
\end{table}

\section{Related work}

Reproducibility and replicability have long been key elements of the scientific method, but have been gaining renewed prominence recently across a number of disciplines with attention being given to a `reproducibility crisis'. 
For example, in pharmaceutical research, as little as 20-25\% of papers were found to be replicable \cite{prinz2011}. The problem has also been recognised in computer science in general \cite{Collberg2016}.
Reproducibility and replicability have been researched for sometime in  Information Retrieval (IR) since the Grid@CLEF pilot track \cite{ferro2009clef}. The aim was to create a `grid of points' where a point defined the performance of a particular IR system using certain pre-processing techniques on a defined dataset. 
\newcite{louridas2012note} looked at reproducibility in Software Engineering after trying to replicate another authors results and concluded with a list of requirements for papers to be reproducible: 
(a) All data related to the paper, 
(b) All code required to reproduce the paper and 
(c) Documentation for the code and data. 
\newcite{fokkens2013offspring} looked at reproducibility in WordNet similarity and Named Entity Recognition finding five key aspects that cause experimental variation and therefore need to be clearly stated: 
(a) pre-processing, (b) experimental setup, (c) versioning, (d) system output, (e) system variation. 
In Twitter sentiment analysis, \newcite{sygkounas2016replication} stated the need for using the same library versions and datasets when replicating work.

Different methods of releasing datasets and code have been suggested. \newcite{ferro2009clef} defined a framework (CIRCO) that enforces a pre-processing pipeline where data can be extracted at each stage therefore facilitating a validation step. They stated a mechanism for storing results, dataset and pre-processed data\footnote{\url{http://direct.dei.unipd.it/}}. \newcite{louridas2012note} suggested the use of a virtual machine alongside papers to bundle the data and code together, while most state the advantages of releasing source code ~\cite{fokkens2013offspring,potthast2016wrote,sygkounas2016replication}.
The act of reproducing or replicating results is not just for validating research but to also show how it can be improved. \newcite{ferro2016clef} followed up their initial research and were able to analyse which pre-processing techniques were important on a French monolingual dataset and how the different techniques affected each other given an IR system. \newcite{fokkens2013offspring} showed how changes in the five key aspects affected results.

The closest related work to our reproducibility study is that of \newcite{E17-4003} which they replicate three different syntactic based aspect extraction methods. They found that parameter tuning was very important however using different pre-processing pipelines such as Stanford's CoreNLP did not have a consistent effect on the results. They found that the methods stated in the original papers are not detailed enough to replicate the study as evidenced by their large results differential.

\newcite{dashtipour2016multilingual} undertook a replication study in sentiment prediction, however this was at the document level and on different datasets and languages to the originals. 
In other areas of (aspect-based) sentiment analysis, releasing code for published systems has not been a high priority, e.g. in SemEval 2016 task 5 \cite{pontiki_2016} only 1 out of 21 papers released their source code. In IR, specific reproducible research tracks have been created\footnote{\url{http://ecir2016.dei.unipd.it/call_for_papers.html}} and we are pleased to see the same happening at COLING 2018\footnote{\url{http://coling2018.org/}}. 


Turning now to the focus of our investigations, Target Dependent sentiment analysis (TDSA) research \cite{nasukawa_2003} arose as an extension to the coarse grained analysis of document level sentiment analysis \cite{pang_2002,turney_2002}. Since its inception, papers have applied different methods such as feature based \cite{kiritchenko_2014}, Recursive Neural Networks (RecNN) \cite{dong_2014}, Recurrent Neural Networks (RNN) \cite{Tang_2016_tdlstm}, attention applied to RNN \cite{wang_2016,chen_2017,tay_2017}, Neural Pooling (NP) \cite{vo_2015,wang_2017}, RNN combined with NP \cite{zhang_2016}, and attention based neural networks \cite{Tang_2016_mem}. 
Others have tackled TDSA as a joint task with target extraction, thus treating it as a sequence labelling problem. \newcite{mitchell_2013} carried out this task using Conditional Random Fields (CRF), and this work was then extended using a neural CRF \cite{zhang_2015}. Both approaches found that combining the two tasks did not improve results compared to treating the two tasks separately, apart from when considering POS and NEG when the joint task performs better.
Finally, \newcite{Marrese_Taylor_2017} created an attention RNN for this task which was evaluated on two very different datasets containing written and spoken (video-based) reviews where the domain adaptation between the two shows some promise. 
Overall, within the field of sentiment analysis there are other granularities such as sentence level \cite{socher_2013}, topic \cite{augenstein_2018}, and aspect \cite{wang_2016,tay_2017}. Aspect-level sentiment analysis relates to identifying the sentiment of (potentially multiple) topics in the same text although this can be seen as a similar task to TDSA. However the clear distinction between aspect and TDSA is that TDSA requires the target to be mentioned in the text itself while aspect-level employs a conceptual category with potentially multiple related instantiations in the text.

\newcite{Tang_2016_tdlstm} created a Target Dependent LSTM (TDLSTM) which encompassed two LSTMs either side of the target word, then improved the model by concatenating the target vector to the input embeddings to create a Target Connected LSTM (TCLSTM). Adding attention has become very popular recently. \newcite{Tang_2016_mem} showed the speed and accuracy improvements of using multiple attention layers only over LSTM based methods, however they found that it could not model complex sentences e.g. negations. \newcite{liu_2017} showed that adding attention to a Bi-directional LSTM (BLSTM) improves the results as it takes the importance of each word into account with respect to the target. \newcite{chen_2017} also combined a BLSTM and attention, however they used multiple attention layers and combined the results using a Gated Recurrent Unit (GRU) which they called Recurrent Attention on Memory (RAM), and they found this method to allow models to better understand more complex sentiment for each comparison.
\newcite{vo_2015} used neural pooling features e.g. max, min, etc of the word embeddings of the left and right context of the target word, the target itself, and the whole Tweet. They inputted the features into a linear SVM, and showed the importance of using the left and right context for the first time. They found in their study that using a combination of Word2Vec embeddings and sentiment embeddings \cite{tang_2014} performed best alongside using sentiment lexicons to filter the embedding space.
Other studies have adopted more linguistic approaches. \newcite{wang_2017} extended the work of \newcite{vo_2015} by using the dependency linked words from the target. \newcite{dong_2014} used the dependency tree to create a Recursive Neural Network (RecNN) inspired by \newcite{socher_2013} but compared to \newcite{socher_2013} they also utilised the dependency tags to create an Adaptive RecNN (ARecNN).


Critically, the methods reported above have not been applied to the same datasets, therefore a true comparative evaluation between the different methods is somewhat difficult. This has serious implications for generalisability of methods. We correct that limitation in our study.
There are two papers taking a similar approach to our work in terms of generalisability although they do not combine them with the reproduction issues that we highlight. First, \newcite{chen_2017} compared results across SemEval's laptop and restaurant reviews in English \cite{pontiki_2014}, a Twitter dataset \cite{dong_2014} and their own Chinese news comments dataset. They did perform a comparison across different languages, domains, corpora types, and different methods; SVM with features \cite{kiritchenko_2014}, Rec-NN \cite{dong_2014}, TDLSTM \cite{Tang_2016_tdlstm}, Memory Neural Network (MNet) \cite{Tang_2016_mem} and their own attention method. However, the Chinese dataset was not released, and the methods were not compared across all datasets. 
By contrast, we compare all methods across all datasets, 
using techniques that are not just from the Recurrent Neural Network (RNN) family.
A second paper, by \newcite{BarnesW17-5202} compares seven approaches to (document and sentence level) sentiment analysis on six benchmark datasets, but does not systematically explore reproduction issues as we do in our paper.




\section{Datasets used in our experiments}
\label{sec:datasets}
We are evaluating our models over six different English datasets deliberately chosen to represent a range of domains, types and mediums. As highlighted above, previous papers tend to only carry out evaluations on one or two datasets which limits the generalisability of their results. 
In this paper, we do not consider the quality or inter-annotator agreement levels of these datasets but it has been noted that some datasets may have issues here. For example, \newcite{pavlopoulos_2014} point out that the \newcite{hu_2004} dataset does not state their inter-annotator agreement scores nor do they have aspect terms that express neutral opinion.

We only use a subset of the English datasets available. For two reasons. First, the time it takes to write parsers and run the models. Second, we only used datasets that contain three distinct sentiments (\newcite{wilson_2008} only has two).
From the datasets we have used, we have only had issue with parsing \newcite{wang_2017} where the annotations for the first set of the data contains the target span but the second set does not. Thus making it impossible to use the second set of annotation and forcing us to only use a subset of the dataset. An as example of this:
``Got rid of bureaucrats `and we put that money, into 9000 more doctors and nurses'... to turn the doctors into bureaucrats\#BattleForNumber10'' in that Tweet `bureaucrats' was annotated as negative but it does not state if it was the first or second instance of `bureaucrats' since it does not use target spans.
As we can see from table \ref{table:datasets_stats}, generally the social media datasets (Twitter and YouTube) contain more targets per sentence with the exception of \newcite{dong_2014} and \newcite{mitchell_2013}. The only dataset that has a small difference between the number of unique sentiments per sentence is the \newcite{wang_2017} election dataset.

Lastly we create training and test splits for the YouTuBean \cite{Marrese_Taylor_2017} and Mitchell \cite{mitchell_2013} datasets as they were both evaluated originally using cross validation. These splits are reproducible using the code that we are open sourcing.


\begin{table}[!h]
\centering
\begin{tabular}{|p{0.11\textwidth}|p{0.04\textwidth}|p{0.04\textwidth}|p{0.05\textwidth}|p{0.08\textwidth}|p{0.05\textwidth}|p{0.06\textwidth}|p{0.06\textwidth}|p{0.055\textwidth}|p{0.05\textwidth}|p{0.04\textwidth}|}
\hline
Dataset & DO & T & Size & M & ATS & Uniq & AVG Len & S1 & S2 & S3\\
 \hline
 \hline
SemEval 14 L& L & RE & 2951 & W & 1.58 & 1295 & 18.57 & 81.09 & 17.62 & 1.29 \\
\hline
SemEval 14 R& R & RE & 4722 & W & 1.83 & 1630 & 17.25 & 75.26 & 22.94 & 1.80 \\
\hline
Mitchel & G & S & 3288 & W & 1.22 & 2507 & 18.02 & 90.48 & 9.43 & 0.09 \\
\hline
Dong Twitter& G & S & 6940 & W & 1.00 & 145 & 17.37 & 100.00 & 0.00 & 0.00 \\
\hline
Election Twitter& P & S & 11899 & W & 2.94 & 2190 & 21.68 & 44.50 & 46.72 & 8.78 \\
\hline
YouTuBean& MP & RE/S& 798 & SP & 2.07 & 522 & 22.53 & 81.45 & 18.17 & 0.38 \\
\hline
\hline
\multicolumn{11}{|p{15cm}|}{L=Laptop, R=Restaurant, P=Politics, MP=Mobile Phones, G=General, T=Type, RE=Review, S=Social Media, ATS=Average targets per sentence, Uniq=No. unique targets, AVG len=Average sentence length per target, S1=1 distinct sentiment per sentence, S2=2 distinct sentiments per sentence, S3=3 distinct sentiments per sentence, DO=Domain, M=Medium, W=Written, SP=Spoken}\\
\hline
\end{tabular}
\caption{Dataset Statistics}
\label{table:datasets_stats}
\end{table}

\section{Reproduction studies}
In the following subsections, we present the three different methods that we are reproducing and how their results differ from the original analysis. In all of the experiments below, we lower case all text and tokenise using Twokenizer \cite{gimpel_2011}. This was done as the datasets originate from Twitter and this pre-processing method was to some extent stated in \newcite{vo_2015} and assumed to be used across the others as they do not explicitly state how they pre-process in the papers.


\subsection{Reproduction of \newcite{vo_2015}}
\newcite{vo_2015} created the first NP method for TDSA. It takes the word vectors of the left, right, target word, and full tweet/sentence/text contexts and performs max, min, average, standard deviation, and product pooling over these contexts to create a feature vector as input to the Support Vector Machine (SVM) classifier. This feature vector is in affect an automatic feature extractor. They created four different methods: 1. \textbf{Target-ind} uses only the full tweet context, 2. \textbf{Target-dep-} uses left, right, and target contexts, 3. \textbf{Target-dep} Uses both features of \textbf{Target-ind} and \textbf{Target-dep-}, and 4. \textbf{Target-dep+} Uses the features of \textbf{Target-dep} and adds two additional contexts left and right sentiment (LS \& RS) contexts where only the words within a specified lexicon are kept and the rest of the words are zero vectors. All of their experiments are performed on \newcite{dong_2014} Twitter data set. For each of the experiments below we used the following configurations unless otherwise stated: we performed 5 fold stratified cross validation, features are scaled using Max Min scaling before inputting into the SVM, and used the respective C-Values for the SVM stated in the paper for each of the models.

One major difficulty with the description of the method in the paper and re-implementation is handling the same target multiple appearances issue as originally raised by \newcite{wang_2017}. As the method requires context with regards to the target word, if there is more than one appearance of the target word then the method does not specify which to use. We therefore took the approach of \newcite{wang_2017} and found all of the features for each appearance and performed median pooling over features. This change could explain the subtle differences between the results we report and those of the original paper.

\subsubsection{Sentiment Lexicons}
%
%
\newcite{vo_2015} used three different sentiment lexicons: MPQA\footnote{\url{http://mpqa.cs.pitt.edu/lexicons/subj_lexicon/}} \cite{wilson_2005}, NRC\footnote{\url{http://saifmohammad.com/WebPages/NRC-Emotion-Lexicon.htm}} \cite{mohammad_2010}, and HL\footnote{\url{https://www.cs.uic.edu/~liub/FBS/sentiment-analysis.html#lexicon}} \cite{hu_2004}. We found a small difference in word counts between their reported statistics for the MPQA lexicons and those we performed ourselves, as can be seen in the bold numbers in table \ref{table:senti_lexicon_stats}. 
Originally, we assumed that a word can only occur in one sentiment class within the same lexicon, and this resulted in differing counts for all lexicons. This distinction is not clearly documented in the paper or code. However, our assumption turned out to be incorrect, giving a further illustration of why detailed descriptions and documentation of all decisions is important.
%
We ran the same experiment as \newcite{vo_2015} to show the effectiveness of sentiment lexicons the results can be seen in table \ref{table:senti_target_results}. We can clearly see there are some difference not just with the accuracy scores but the rank of the sentiment lexicons. We found just using HL was best and MPQA does help performance compared to the \textbf{Target-dep} baseline which differs to \newcite{vo_2015} findings. Since we found that using just HL performed best, the rest of the results will apply the \textbf{Target-dep+} method using HL and using HL \& MPQA to show the affect of using the lexicon that both we and \newcite{vo_2015} found best.  
\begin{table}[!h]
\centering
\begin{tabular}{|l|c|c|c|c|c|c|}
\hline
 &  \multicolumn{6}{c|}{Word Counts}  \\
\hline
 & \multicolumn{2}{c|}{Original} & \multicolumn{4}{c|}{Reproduction} \\
\hline
\hline
Lexicons & Positive & Negative & Positive & Positive Lowered & Negative & Negative Lowered \\
\hline
MPQA & 2289 & 4114 & \textbf{2298} & \textbf{2298} & \textbf{4148} & \textbf{4148} \\
\hline
HL & 2003 & 4780 & 2003 & 2003 & 4780 & 4780 \\
\hline
NRC & 2231 & 3243 & 2231 & 2231 & 3243 & 3243 \\
\hline
MPQA \& HL & 2706 & 5069 & \textbf{2725} & \textbf{2725} & \textbf{5080} & \textbf{5076} \\
\hline
All three & 3940 & 6490 & \textbf{4016} & \textbf{4016} & \textbf{6530} & \textbf{6526} \\
\hline
\end{tabular}
\caption{Sentiment lexicon statistics comparison}
\label{table:senti_lexicon_stats}
\end{table}

\begin{table}[h!]
\centering
\begin{tabular}{|l|c|c|}
\hline
 & \multicolumn{2}{|c|}{Results (Accuracy \%)} \\
\hline
Sentiment Lexicon & Original & Reproduction \\
\hline
\hline
Target-dep & 65.72 & 66.81  \\
\hline
Target-dep+: NRC & 66.05 & 67.13 \\
\hline
Target-dep+: HL & 67.24 & \textbf{68.61} \\
\hline
Target-dep+: MPQA & 65.56 & 66.81 \\
\hline
Target-dep+: MPQA \& HL & \textbf{67.40} & 68.37 \\
\hline
Target-dep+: All three & 67.30 & 68.23 \\
\hline
\end{tabular}
\caption{Effectiveness of Sentiment Lexicons}
\label{table:senti_target_results}
\end{table}

\subsubsection{Using different word vectors}
\begin{figure}[!h]
\centering
\includegraphics[scale=0.55]{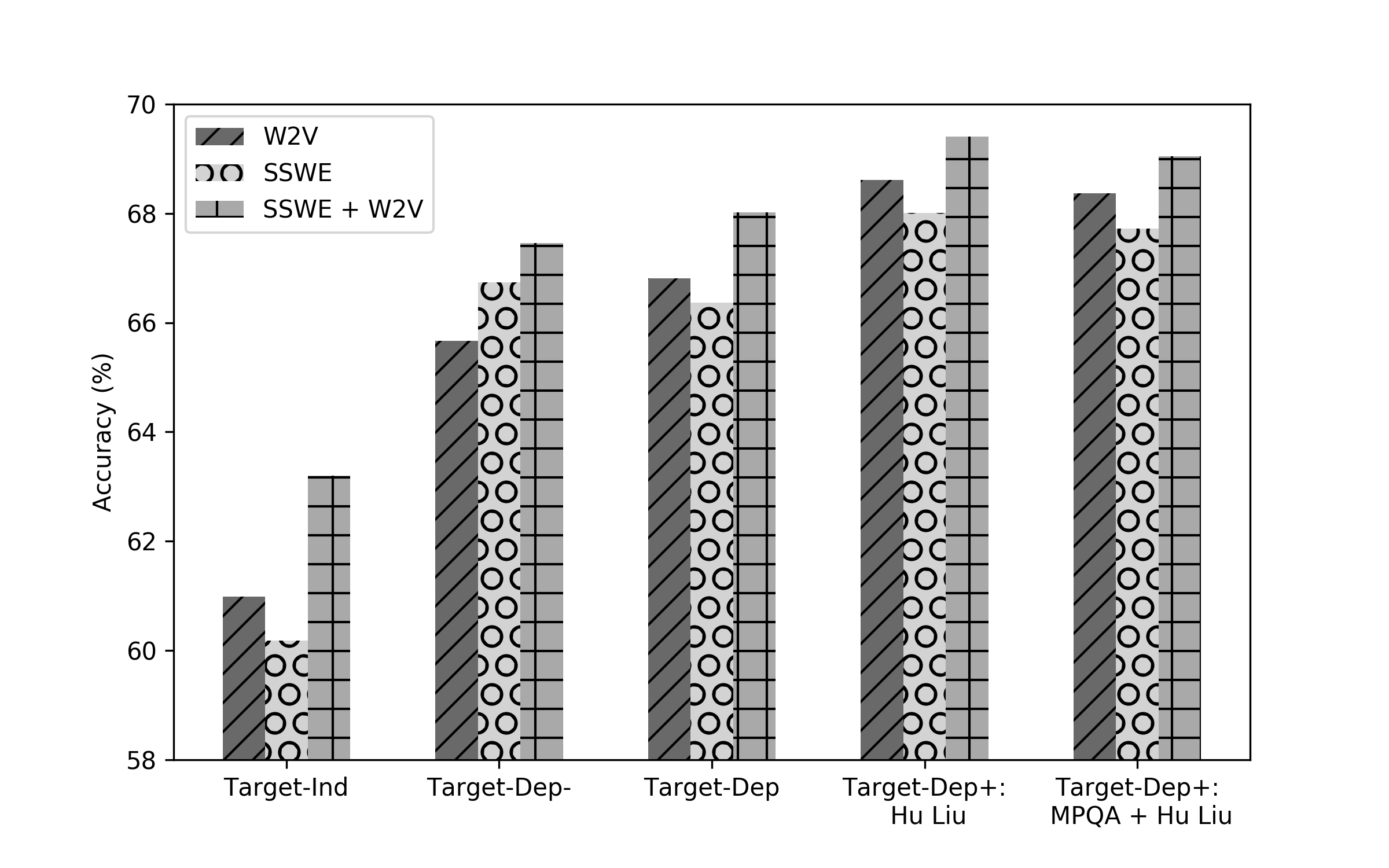}
\caption{Effectiveness of word embedding}
\label{embeddings_target_results}
\end{figure}

The original authors tested their methods using three different word vectors: 1. Word2Vec trained by \newcite{vo_2015} on 5 million Tweets containing emoticons (W2V), 2. Sentiment Specific Word Embedding (SSWE) from \newcite{tang_2014}, and 3. W2V and SSWE combined. Neither of these word embeddings are available from the original authors as \newcite{vo_2015} never released the embeddings and the link to \newcite{tang_2014} embeddings no longer works\footnote{\url{http://ir.hit.edu.cn/~dytang/}}. However, the embeddings were released through \newcite{wang_2017} code base\footnote{\url{https://github.com/bluemonk482/tdparse}} following requesting of the code from \newcite{vo_2015}. Figure \ref{embeddings_target_results}
shows the results of the different word embeddings across the different methods. 
The main finding we see is that SSWE by themselves are not as informative as W2V vectors which is different to the findings of \newcite{vo_2015}. However we agree that combining the two vectors is beneficial and that the rank of methods is the same in our observations.


\subsubsection{Scaling and Final Model comparison}
\begin{figure}[!h]
\centering
\includegraphics[scale=0.55]{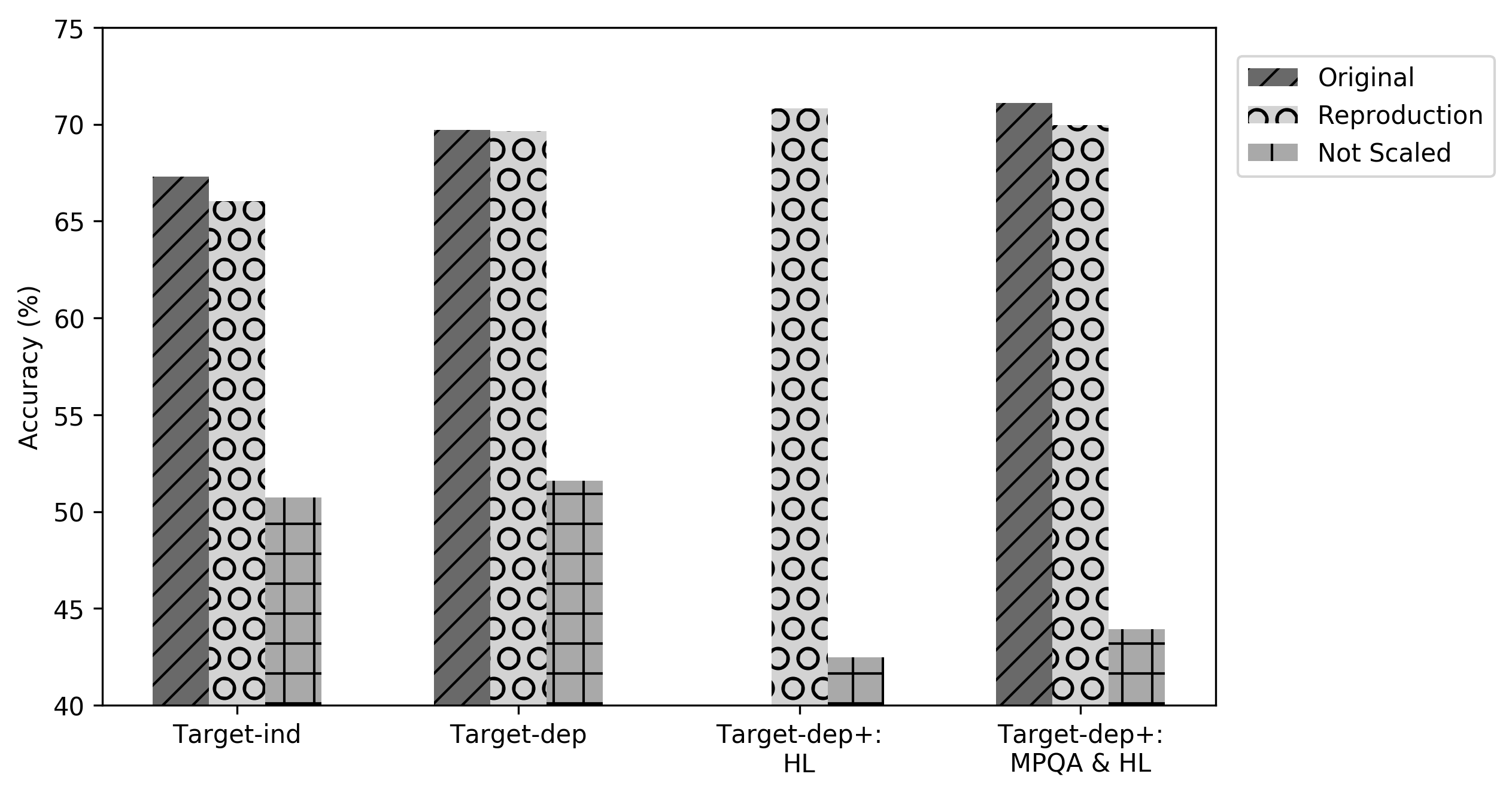}
\caption{Target Dependent Final Results}
\label{figure:target-dep-results}
\end{figure}

We test all of the methods on the test data set of \newcite{dong_2014} and show the difference between the original and reproduced models in figure \ref{figure:target-dep-results}. Finally, we show the effect of scaling using Max Min and not scaling the data. 


As stated before, we have been using Max Min scaling on the NP features, however \newcite{vo_2015} did not mention scaling in their paper. The library they were using, LibLinear \cite{fan_2008}, suggests in its practical guide \cite{hsu_2003} to scale each feature to [0, 1] but this was not re-iterated by \newcite{vo_2015}. We are using scikit-learn's \cite{scikit-learn} LinearSVC which is a wrapper of LibLinear, hence making it appropriate to use here. As can be seen in figure \ref{figure:target-dep-results}, not scaling can affect the results by around one-third. 

\subsection{Reproduction of \newcite{wang_2017}}
\begin{figure}[h!]
\centering
\includegraphics[scale=0.55]{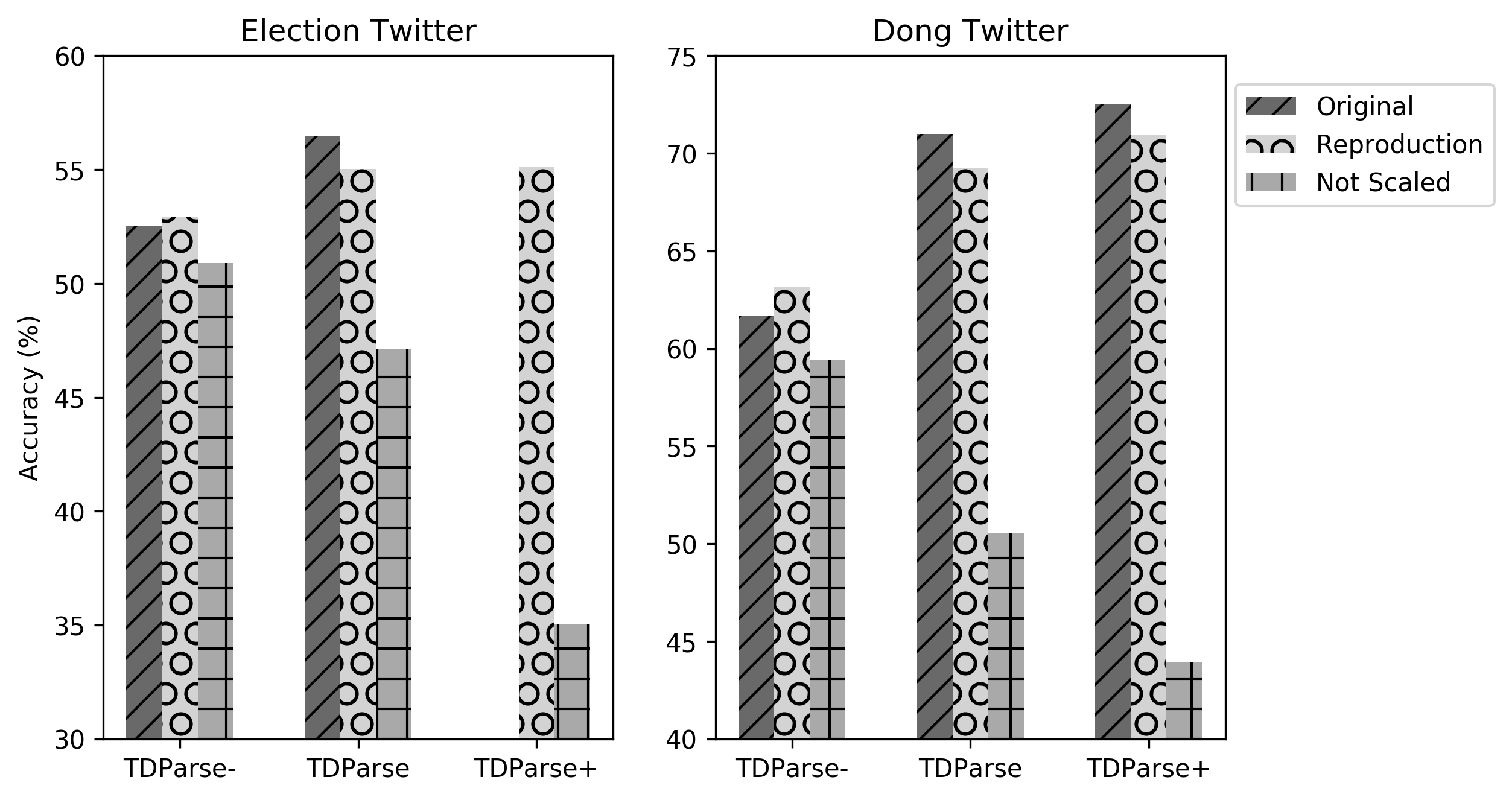}
\caption{TDParse Final Results}
\label{figure:tdparse-results}
\end{figure}

\newcite{wang_2017} extended the NP work of \newcite{vo_2015} and instead of using the full tweet/sentence/text contexts they used the full dependency graph of the target word. Thus, they created three different methods: 1. \textbf{TDParse-} uses only the full dependency graph context, 2. \textbf{TDParse} the feature of \textbf{TDParse-} and the left and right contexts, and 3. \textbf{TDParse+} the features of \textbf{TDParse} and LS and RS contexts. The experiments are performed on the \newcite{dong_2014} and \newcite{wang_2017} Twitter datasets where we train and test on the previously specified train and test splits. We also scale our features using Max Min scaling before inputting into the SVM. We used all three sentiment lexicons as in the original paper, and we found the C-Value by performing 5 fold stratified cross validation on the training datasets. The results of these experiments can be seen in figure \ref{figure:tdparse-results}\footnote{For the Election Twitter dataset TDParse+ result were never reported in the original paper.}. As found with the results of \newcite{vo_2015} replication, scaling is very important but is typically overlooked when reporting.


\subsection{Reproduction of \newcite{Tang_2016_tdlstm}}
\newcite{Tang_2016_tdlstm} was the first to use LSTMs specifically for TDSA. They created three different models: 1. \textbf{LSTM} a standard LSTM that runs over the length of the sentence and takes no target information into account, 2. \textbf{TDLSTM} runs two LSTMs, one over the left and the other over the right context of the target word and concatenates the output of the two, and 3. \textbf{TCLSTM} same as the \textbf{TDLSTM} method but each input word vector is concatenated with vector of the target word. All of the methods outputs are fed into a softmax activation function. The experiments are performed on the \newcite{dong_2014} dataset where we train and test on the specified splits. For the LSTMs we initialised the weights using uniform distribution U(−0.003, 0.003), used Stochastic Gradient Descent (SGD) a learning rate of 0.01, cross entropy loss, padded and truncated sequence to the length of the maximum sequence in the training dataset as stated in the original paper, and we did not ``set the clipping threshold of softmax layer as 200'' \cite{Tang_2016_tdlstm} as we were unsure what this meant. With regards to the number of epochs trained, we used early stopping with a patience of 10 and allowed 300 epochs. Within their experiments they used SSWE \cite{tang_2014} and Glove Twitter vectors\footnote{\url{https://nlp.stanford.edu/projects/glove/}} \cite{pennington_2014}.

\begin{figure}
\begin{floatrow}

\capbtabbox{%
\centering
\begin{tabular}{|l|K|K|K|}
\hline
 & \multicolumn{3}{c|}{Macro F1} \\
\hline
Methods& O & R (Max) & R (Mean) \\
\hline
\hline
LSTM & 64.70 & 64.34 & 60.69 \\
\hline
TDLSTM & 69.00 & 67.04 & 65.63\\
\hline
TCLSTM & 69.50 & 67.66 & 65.23\\
\hline
\hline
\multicolumn{4}{|c|}{O=Original, R=Reproduction}\\
\hline
\end{tabular}
}{%
\caption{LSTM Final Results}
\label{table:tdlstm-results}
}
	

\ffigbox{%
\centering
  \includegraphics[scale=0.5]{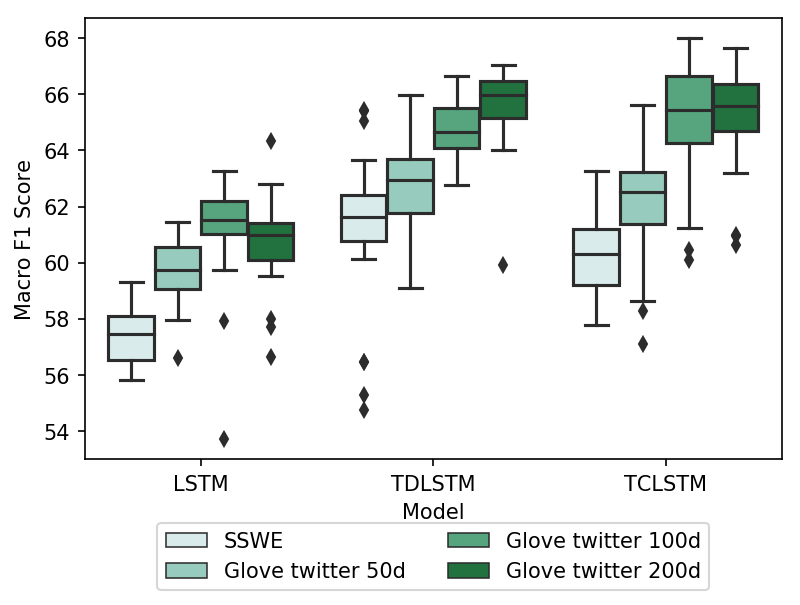}
}{%
\caption{Distribution of the LSTM results}
\label{figure:lstm_dist_results}
}{}

\end{floatrow}
\end{figure}

As the paper being reproduced does not define the number of epochs they trained for, we use early stopping. Thus for early stopping we require to split the training data into train and validation sets to know when to stop. As it has been shown by \newcite{reimers_2017} that the random seed statistically significantly changes the results of experiments we ran each model over each word embedding thirty times, using a different seed value but keeping the same stratified train and validation split, and reported the results on the same test data as the original paper. As can be seen in Figure \ref{figure:lstm_dist_results}, the initial seed value makes a large difference more so for the smaller embeddings. In table \ref{table:tdlstm-results}, we show the difference between our mean and maximum result and the original result for each model using the 200 dimension Glove Twitter vectors. Even though the mean result is quite different from the original the maximum is much closer. Our results generally agree with their results on the ranking of the word vectors and the embeddings.

Overall, we were able to reproduce the results of all three papers. However for the neural network/deep learning approach of \newcite{Tang_2016_tdlstm} we agree with \newcite{reimers_2017} that reporting multiple runs of the system over different seed values is required as the single performance scores can be misleading, which could explain why previous papers obtained different results to the original for the \textbf{TDLSTM} method \cite{chen_2017,tay_2017}. 




\section{Mass Evaluation}
For all of the methods we pre-processed the text by lower casing and tokenising using Twokenizer \cite{gimpel_2011}, and we used all three sentiment lexicons where applicable. We found the best word vectors from SSWE and the common crawl 42B 300 dimension Glove vectors by five fold stratified cross validation for the NP methods and the highest accuracy on the validation set for the LSTM methods. We chose these word vectors as they have very different sizes (50 and 300), also they have been shown to perform well in different text types; SSWE for social media \cite{Tang_2016_tdlstm} and Glove for reviews \cite{chen_2017}. To make the experiments quicker and computationally less expensive, we filtered out all words from the word vectors that did not appear in the train and test datasets, and this is equivalent with respect to word coverage as using all words. Finally we only reported results for the LSTM methods with one seed value and not multiple due to time constraints.

The results of the methods using the best found word vectors on the test sets can be seen in table \ref{table:mass_eval}. We find that the \textbf{TDParse} methods generally perform best but only clearly outperforms the other non-dependency parser methods on the YouTuBean dataset. We hypothesise that this is due to the dataset containing, on average, a deeper constituency tree depth which could be seen as on average more complex sentences. This could be due to it being from the spoken medium compared to the rest of the datasets which are written. Also that using a sentiment lexicon is almost always beneficial, but only by a small amount. Within the LSTM based methods the \textbf{TDLSTM} method generally performs the best indicating that the extra target information that the \textbf{TCLSTM} method contains is not needed, but we believe this needs further analysis.

We can conclude that the simpler NP models perform well across domain, type and medium and that even without language specific tools and lexicons they are competitive to the more complex LSTM based methods.

\begin{table}[!h]
\centering
\begin{tabular}{|l|Z|Z|T|V|Z|V|V|}
\hline
Dataset & Target-Dep F1 & Target-Dep+ F1 & TDParse F1 & TDParse+ F1 & LSTM F1 & TDLSTM F1 & TCLSTM F1\\
\hline
\hline
Dong Twitter & 65.70 & 65.70 & 66.00 & \textbf{68.10} & 63.60 & 66.09 & 67.14 \\
\hline
Election Twitter & 45.50 & 45.90 & \textbf{46.20} & 44.60 & 38.70 & 43.60 & 42.08 \\
\hline
Mitchell & 40.80 & 42.90 & 40.50 & 50.00 & 47.17 & \textbf{51.16} & 41.03 \\
\hline
SemEval 14 L & 60.00 & 63.70 & 59.60 & \textbf{64.50} & 47.84 & 57.91 & 46.80 \\
\hline
SemEval 14 R & 56.20 & 57.70 & 59.40 & \textbf{61.00} & 46.36 & 57.68 & 55.38 \\
\hline
YouTuBean & 53.10 & 55.60 & \textbf{71.70} & 68.00 & 45.93 & 45.47 & 38.07 \\
\hline
Mean & 53.55 & 55.25 & 57.23 & \textbf{59.37} & 48.27 & 53.65 & 48.42 \\
\hline
\end{tabular}
\caption{Mass Evaluation Results}
\label{table:mass_eval}
\end{table}

				
\section{Discussion and conclusion}
The fast developing subfield of TDSA has so far lacked a large-scale comparative mass evaluation of approaches using different models and datasets. In this paper, we address this generalisability limitation and perform the first direct comparison and reproduction of three different approaches for TDSA.
While carrying out these reproductions, we have noted and described above, the many emerging issues in previous research related to incomplete descriptions of methods and settings, patchy release of code, and lack of comparative evaluations. This is natural in a developing field, but it is crucial for ongoing development within NLP in general that improved repeatability practices are adopted. The practices adopted in our case studies are to reproduce the methods in open source code, adopt only open data, provide format conversion tools to ingest the different data formats, and describe and document all settings via the code and Jupyter Notebooks (released initially in anonymous form at submission time)\footnote{\url{https://github.com/apmoore1/Bella}}. We therefore argue that papers should not consider repeatability (replication or reproduction) or generalisability alone, but these two key tenets of scientific practice should be brought together.

In future work, we aim to extend our reproduction framework further, and extend the comparative evaluation to languages other than English. This will necessitate changes in the framework since we expect that dependency parsers and sentiment lexicons will be unavailable for specific languages. Also we will explore through error analysis in which situations different neural network architectures perform best.


\section*{Acknowledgements}
This research is funded at Lancaster University by an EPSRC Doctoral Training Grant.

\bibliographystyle{acl.bst}
\bibliography{target_sentiment.bib}

\end{document}